\documentclass[a4paper]{article}
\usepackage{graphicx,amssymb}
\begin{document}

\noindent{\large {\bf{Artificial Neural Network-based  error compensation  procedure  for low-cost encoders.}}}
\vskip 0.5cm
\noindent{\large {\bf{ V.K.Dhar, A.K.Tickoo, S.K.Kaul, R.Koul, B.P.Dubey $^\dagger$}}}
%\begin{center}

\vskip 0.3cm
{Astrophysical Sciences Division, \\
$^\dagger$ Electronics and Instruments Services Division, \\
Bhabha Atomic Research Centre, Mumbai-400 085,India.}

\vskip 1.0 cm
%\begin{center}
\noindent{\bf Abstract }:\\
An Artificial Neural Network-based  error compensation method  is proposed   for improving  the accuracy of resolver-based 
16-bit encoders  by  compensating  for  their  respective  systematic   error  profiles.  
The   error compensation procedure,  for a particular  encoder,   involves  obtaining  its error profile  by calibrating it  on a precision rotary table,   training  the neural network  by  using  a  part of  this data  and then  determining  the   corrected  encoder angle   by  subtracting the ANN-predicted  error  from the  measured value of the encoder angle.  Since  it  is  not guaranteed   that   all  the  resolvers   will  have   exactly  similar  error  profiles  because  of  the  inherent  differences  in their  construction  on  a  micro  scale,  the ANN   has been   trained  on  one  error profile at a time and  the  corresponding  weight  file  is  then  used only  for  compensating   the  systematic  error  of  this  particular  encoder.   The  systematic  nature  of  the  error profile for  each of the encoders has  also  been  validated  by   repeated  calibration  of the encoders  over a period  of  time  and  it was found  that  the  error profiles  of a  particular  encoder  recorded  at different  epochs   show  near reproducible  behaviour.  
The    ANN-based  error compensation  procedure   has  been   implemented   for 4  encoders by  training 
the  ANN  with  their  respective  error profiles   and  the  results  indicate  that  the  accuracy of  encoders  can  be improved  by nearly an  order of  magnitude  from   quoted   values  of   $\sim$$\pm$ 6 arc-min  to $\sim$$\pm$0.65 arc-min  when  their  corresponding   ANN-generated   weight  files  are  used  for  determining  the  corrected  encoder angle.   \newline
\\
%\end{center}
   
\vskip 0.3cm
{\it Key Words : \hspace{0.01cm} Resolver based encoder, Error compensation, Artificial Neural Network. \\}
%\end{abstract} 

%\end{center}
\noindent{\bf{email: veer@barc.gov.in}}

\newpage
%-----------------------------------------------------------------------------------------------------------------------------------------------------------------

\noindent {\bf 1. Introduction :}
\vskip 0.5cm
%----------------------------------------------------------------------------
A resolver [1,2] is an electromechanical device which  converts shaft angle to an absolute analog signal.  The  construction of a  resolver  is  similar to that  of an AC electric motor with two sets of phase windings acting together as a variable phase transformer, whose output analog voltage represents the input shaft angle uniquely.  On excitation by a signal, one of the sense windings of the resolver develops a voltage proportional to the sine of the rotor displacement angle  and  the other develops a voltage proportional to the cosine of this angle. A complete  encoder system is built  by coupling a resolver  to an excitation source and phase angle monitoring circuit known as resolver-to-digital (R/D) converter [3]. The conversion of the resolver signal to position is achieved by using a tracking loop which monitors the phase lag between the actual and measured position.  
Resolvers  have been available  for  a  few  decades  in various forms as part of electromechanical  shaft angle  position measurement  systems. 
They  are robust, reliable and have a long life even while working in severe and harsh  environments.  High reliability and lower price compared to optical encoders makes  resolver-based encoders an  ideal choice for applications involving moderate accuracies of  $\sim$10 arc-min.
\par
Keeping  in view the fact  that,  in reality,   no resolver  can generate  ideal sinusoidal  signals  because of the manufacturing  tolerances, it is obvious  that  low-cost resolvers   will  be   available with  only modest  accuracies. The non-ideal characteristics  of  a resolver have  been discussed  thoroughly in the seminal work of Hanselman [4,5]  and  it is   now very  well established   that  the  most  important  contributors to the error profile budget  are the following: (i) Amplitude imbalance, (ii) Quadrature error, (iii) Inductive harmonics, and (iv) Excitation signal distortion. The  non-ideal  characteristics  of a resolver  arise  because of the finite precision with which a resolver can be mechanically constructed  and the  non-ideal  nature  of  the excitation signal.  While  low-cost resolvers  are available with accuracy values of about 
$\pm$10 arc-min, improvement in their accuracy  requires maintaining   stringent tolerance limits in their manufacture, which  can  add to the cost considerably.  
\par
The easiest way to reduce or compensate position error  due to mechanical misalignments and imperfections  in the signal outputs of a resolver  is to calibrate it against a higher accuracy sensor  so that the integrated  systematic error  profile of the resolver  along with its  R/D converter  can be measured  and    then  compensated by using a  suitable procedure.  The results of  our software-based  compensation  work, using   the  look up table  approach [6] and Fourier-series  based  method [7]  also    confirm that the  accuracy of  low-cost, resolver based 16-bit encoders  can  indeed be improved  from   quoted  accuracy   $\sim$$\pm$ 6 arc-min  to $\sim\pm$ 0.65 arc-min.  The  main aim of the present work  is  to  study  the feasibility of using  Artificial Neural Network  for predicting  the  error profile  of   encoders   so that  appropriate  correction can be applied  to  the measured  values of the  encoder angle for determining  the corrected  encoder angle.   Since  all  the  resolvers  will not have   exactly  similar  error  profiles  becaue  of  the  inherent  differences  in their  construction,   the  ANN   needs  to be    trained  on  one  error profile at a time.  Hence,   corresponding  to  each  encoder,  there  is  a  unique  ANN-generated  weight  file  which  one  has  to  use  for  compensating   the  systematic  error.   
%-------------------------------------------------------------------------------------------------------------

\vskip 0.5cm
\noindent{\bf{ Summary  of  some  compensation applications}}
\vskip 0.5cm

Precision motion systems  play an important and direct role in the industry. In these systems involving automated positioning machines and other machine tools, the relative position errors between the machine and the work piece directly affect the quality of the final product or the process concerned.  No matter how well the machine may be designed and manufactured, these error sources are inevitable in any motion system  and  hence  there  is an inherent limit to the achievable accuracy on these machines. While, careful design and precise mechanical  construction  can reduce these errors,  error modeling and compensation is a highly  viable option in improving the system performance further  at a much reduced cost. The basic motivation behind the compensation is to   measure the magnitude of inaccuracy and compensate it through various compensation methods.  As long as the errors are systematic, repeatable and measurable, they can be compensated by using appropriate techniques. 

Conventionally, compensation methods utilize mechanical correctors, leadscrew correction etc. for improving accuracy. However, these devices increase the complexity of the machine and over a period of time, due to wear and tear,  degrade the error compensation. More so, even these corrective components have to be monitored, serviced, calibrated or even replaced on a regular basis resulting in higher costs and downtime. Software based error compensation methods include use of fuzzy error interpolation techniques, neural-based approaches, genetic algorithms, finite element analysis and   Multi-variant Regression analysis.   All of these methods work on a static geometric model of the machine errors, which are obtained (or derived) from measurements of the machine with reference standards at various predetermined points.  With the present day error compensation methods, the conventional limits on accuracy of machine tools can be overcome significantly [8].
A great deal of work in the recent past has gone in identification and elimination of sources of error in machine systems.  The machine manufacturers are now able to achieve better accuracies on account of improved design followed  by  an  appropriate   error  compensation methodology.   

The use of ANN methodology in error compensation has been reported in many studies for various applications.  A feed-forward neural network  employing  11:15:5 model (11 inputs, 15 hidden nodes and 5 outputs) has  been  used to   predict  five  thermally induced spindle errors  with  an  accuracy  better than  $\pm$15$\%$ [9]. 
Heng-Ming Tai  et al [10] have used an ANN based algorithm for tracking control of industrial drive systems.  Although several tracking control techniques like sliding mode, variable structure, self-tuning and model reference adaptive controls have been used but ANN based tracking controls have been found to be ideally suited for this application.     Using   a   backpropagation based ANN algorithm,   the  authors  of the above  work   claim  that  the ANN based controller  can  achieve   real time tracking of any arbitrary prescribed trajectory with a higher degree of accuracy.  The idea of using ANN based models for physical systems has also  been  explored  by several other researchers  [11-17] to understand the characteristics of inverse dynamics of controlled systems .  
An  ANN based approach   has  been  used    for  error compensation  of machine tools also [18].  Important sources of error in CNC machine tools are the geometric motions of the individual machine elements along with the thermal errors which cause these geometric errors to vary over time.  A three-layer feedforward network with 2 inputs   (corresponding to x and z coordinates in the workspace), 2 outputs (corresponding to position error components) and 12 nodes in the hidden layer has been employed by them for machine error compensation.  The results of their study demonstrate that substantial improvements in positioning accuracy are obtained through the use of ANN methodology.   It  is also emphasized    in this   work   that    the algorithm would however  fail  if tested on another machine with different error characteristics.   
K.K.Tan  et al [19]  have  applied ANN based  compensation for geometrical errors for coordinate measuring machines (CMM) to minimize the position error between  end-effector and  workpiece. 
   
%-------------------------------------------------------------------------------------------------------
\vskip 0.5cm
\noindent{\bf{ Measurement  of encoder  error profiles}}
\vskip 0.5cm
Some  of  the  main specifications of  the  single turn 16-bit encoders ,  which are  used in the present  study,  are the following [20]:  Make --- CCC,USA; Model -- HR 90-11;  Size --- NEMA 12; Resolution --- $\sim$0.33 arc-min (16-bit); Accuracy  --- $\pm$6.0 arc-min.
Eight such  encoders  are  being  presently  used in  the  4-element  TACTIC  (TeV Atmospheric Cherenkov Telescope with Imaging Camera) array  of atmospheric  Cherenkov telescopes [21]. Each telescope  of the array  uses two  encoders  for  monitoring its zenith and azimuth angle so that  the drive control software  [22]  can  be operated  in a closed-loop configuration  for tracking  a celestial  source.  The  calibration of the  encoders  was carried out  with a  Rotary Table  of a  Coordinate  Measuring machine  ( Make: Zeiss, Germany; Model : RT 05 -300). The  Rotary Table   has a resolution of $\sim$0.5 arc seconds and an accuracy of $\sim$2 arc seconds [23].   The calibration procedure involves  rotating  the resolver shaft  in  a 2$^\circ$ step  and  recording the decoded resolver angle value alongwith  the  corresponding Rotary Table angle value.   The calibration data  of a particular encoder, which  is needed  for training  the  network,   thus  consists  of a data file of  180   values of the  decoded  resolver angles  and  the corresponding  Rotary Table angles. The  decoded  resolver angles  were  recorded  at Rotary Table angle values of   0$^\circ$, 2$^\circ$, 4$^\circ$, 6$^\circ$, .......  , 358$^\circ$.    
Although the overall error profile  of a encoder system is expected to be  dominated   by  the  non-ideal characteristics  of the resolver, the  error contribution  from  the decoder (comprising   the  excitation  signal source,  R/D converter and  angle read out circuitry) cannot  be completely ignored,   more so,  when the  excitation  signal  source itself  is  contained in the decoder.   Since  using  a single resolver with  two different  decoder  units   indicated   that for  the  same  resolver angle (as measured by the  Rotary Table) there  can  be a  mismatch of  $\sim$2 arc-min  between the  two decoders, we have treated  a particular resolver decoder combination as a single  unit  and calibrated  them together as a single  entity.  Calibration of  the  encoders  performed in this manner and maintaining a particular  resolver decoder combination throughout its use   obviates  recording of  the resolver and the decoder  error profiles individually.  The  systematic  nature  of  the  error profile for  each of the encoder was  also   validated  by   repeated  calibration  of the encoders  over a period  of  time  and  it was found  that  the error profiles  of a  particular  encoder  recorded  at different  epochs  show  near reproducible  behaviour, 
to within the limits of the decoder  resolution.  
In order to check the interpolation  capability of the neural network, an independent test  data sample  was also  taken  for  the 4  encoders.   This data was taken using a  step size  of  2$^\circ$  at Rotary Table angle values of 
1$^\circ$, 3$^\circ$, 5$^\circ$, 7$^\circ$, .......  , 359$^\circ$.  The test data sample,  for a particular encoder,  thus consists  of  another  data file of 180 values  of  Rotary Table  and  the corresponding  encoder angle.  
\par
A representative  example  of the  measured error profiles  for  4 encoders is  shown  in  Fig.1.  The data presented  in  this  figure  uses  both the training  data file  and  the test  data file.  The  error  at a particular  encoder  angle ( $\theta_{ENC}$ )  is  evaluated  by calculating  $\theta_{ENC}$- $\theta_{TAB}$, where  $\theta_{ENC}$  represents the  encoder angle  and  $\theta_{TAB}$ is the corresponding angle recorded by the Rotary Table.   It is evident from this figure  that  the error  profiles  exhibit   widely different patterns.   Thus    having  several  resolvers   of  the  same  tolerance  value   and   from  the  same   batch   does  not necessarily mean  that  the  individual  construction  and   behavior (i.e error  profiles) are  also exactly   similar on  a  micro scale.  Therefore, one   can  not   use  the calibration  data  of  one  resolver   for  compensating   the  systematic  error of  another  resolver.   The mean absolute error (MSE) and RMS error  for the 4 encoders  are shown in Table 1 below. 
%----------------------------------------------------------------------------------------
\begin{table}[h]
\begin{center}
\begin{tabular}{|c|c|c|}
\hline
$ENCODER$  &   $MAE  $     &           $RMS$    \\ 
$ No. $      &   $(arc-min)$   &      $(arc-min)$   \\ 
\hline
$1 $ & $1.33$ & $1.48$  \\ 
\hline
$2$& $0.55$ &  $0.65$  \\ 
\hline
$3$  & $1.09$ &  $1.35$  \\     
\hline
$4 $  &  $1.78$ &  $1.31$ \\
\hline
\end{tabular}
\end{center}
\caption{\it{ Mean Absolute Error (MAE) and the RMS error for the 4 encoders}} 
\end{table}
The  minimum  and  maximum errors  of the encoders which are  observed to be between  $\sim$-4.42 arc-min ( for  Encoder$\#4$)  to   $\sim$3.28 arc-min ( for  Encoder$\#3$)  are quite consistent  with  the tolerance  range  of  $\sim$$\pm$ 6 arc-min  quoted by the manufacturer. 
%---------------------------------------------------------------------------------------------------
%-----Fig.1 --  Error  profile  ----------------------------------------------------------------
\begin{figure}[h]
\centering
\includegraphics*[width=0.8\textwidth,angle=0,clip]{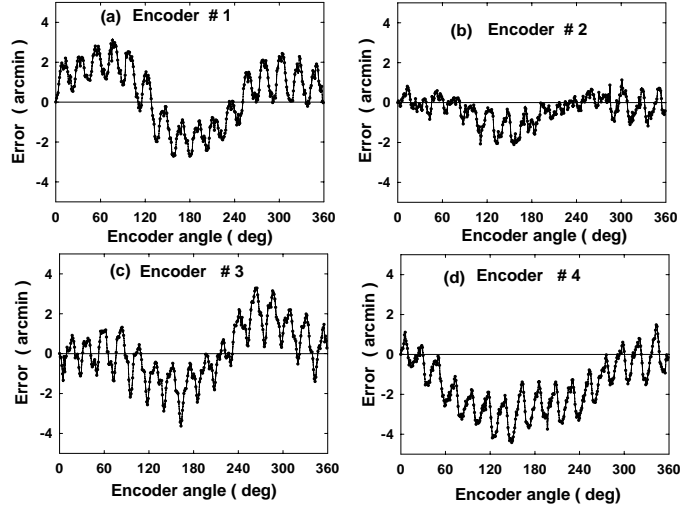}
\caption{\it{ \label{fig ---}(a-d)  The error profiles ( i.e $\theta_{ENC}$- $\theta_{TAB}$)  of  encoders  as a function of the encoder angle ($\theta_{ENC}$). A smoothened spline curve passing through the data points has also been  drawn  in these  figures so that the periodic structure present in the error profiles can be  visualized  easily.} }
\end{figure}

%\end{center}

%---------------------------------------------------------------------------------------------
%-----------------------------------------------------------------------------------------------

\vskip 0.5cm
\noindent{\bf{ Artificial Neural Network  methodology  and  training  of the network}}
\vskip 0.5cm
An Artificial Neural Network (ANN) is an interconnected group of artificial neurons that uses a mathematical model for information processing  to accomplish a variety of tasks like  pattern recognition and classification.  The ability of ANN to handle 
non-linear data interactions, and their robustness in the presence of  high noise levels has encouraged 
their successful use in diverse areas of physics, biology, medicine, agriculture, computer  science 
and  astronomy [24,25].  While the theory and the implementation of ANN has been around for more than 50 years, it is only recently that  it has found wide spread practical applications. This is primarily  due to the advent of high speed, low cost computers that can support the rather computationally intensive requirement of an ANN of any real complexity.
\par
In a feed-forward ANN the network is constructed using layers where all nodes in a given layer are connected to all nodes in a subsequent layer. The network  requires at least two layers, an input layer and an output layer. In addition, the network can include any number of hidden layers with any number of hidden nodes in each layer. The signal from the input vector propagates through the  network layer by layer till the output layer is reached. The output vector  represents the predicted output of the ANN and has a node  for each variable that is being  predicted.  The task of training  the ANN is to find the most appropriate set of weights for each connection  which minimizes  the output error.    All  weighted-inputs are summed  at the neuron node and this summed value is then passed to a transfer (or scaling)  function. 
For  a feed-forward network   with  K input nodes   described  by the input vector (x$_{1}$, x$_{2}$,......),    one hidden  layer with  J nodes  and   I  output nodes,   the output  
F$_{i}$  is given  by the following  equation
%------------------------------------------Equation 1 -----------------------
\begin{equation}
F_{i} = g  \left[ \sum \limits_{j=1}^{J} w_{ij}g  \left(\sum \limits_{k=1}^{K} w_{jk}x_k + \theta_j\right)+\theta_i\right]
\end{equation}
%-------------------------------------------------------------------------------
where    \textit{w}$_{ij}$,\textit{w}$_{jk}$   are the weights, $\theta_i$, $\theta_j$   are  the thresholds   and    g($\ast$) is    the activation function.     
The  training  data sample  is repeatedly  presented to the network in a number of training cycles, and the  adjustment of the free parameters 
(\textit{w}$_{ij}, $, \textit{w}$_{jk}$, $\theta_i$  and $\theta_j$ ) is controlled  by the learning  rate  $\eta$.  The essence of the training process is to iteratively  reduce the error between the predicted value and the target value.   While  the choice of  using  a   particular  error function  is  problem dependent, there  is no well  defined  rule   for  choosing    the most suitable  error function. We have used  the  mean-squared error $MSE$ in this  work  and it is defined as : 
%------------------------------------------Equation 2 -----------------------
\begin{equation}                   
MSE =\frac{1}{PI} {\sum \limits_{p=1}^{P} \sum \limits_{i=1}^{I} \left({D_{p i} - O_{p i}}\right)^2}    
\end{equation}  
%----------------------------------------------------------------
where D$_{pi}$ and O$_{pi}$  are the desired and the observed values   and P  is  number  of training patterns.   The mean-squared error depicts the accuracy of the neural network mapping after a number of training cycles have been implemented. In supervised learning, the correct results (i.e. target values or desired outputs) are known and are given to the neural network  during training so that  it can adjust its weights to  match its outputs to the target values. After training, the  neural network  is tested by giving it only input values, and seeing how close it comes to reproducing  the correct target values. 
\par
Given the inherent power of ANN to effectively handle the multivariate data fitting, we have studied  the feasibility of using it for  predicting  the  error profiles  of   the encoders.  The aim is  to  determine   the  correct encoder  angle   by applying the ANN-predicted  error correction  to  the measured  value of the  encoder angle.  In this work,    we have  used  MATLAB [26]  for  training  and testing   various  ANN  algorithms.  MATLAB  is a numerical  computing environment   which  allows  easy  matrix manipulation,  plotting of functions and data, implementation of algorithms,  creation of user interfaces and interfacing  with  programs  in other  languages.   The  Neural Network  Toolbox  extends  MATLAB  with tools  for  designing, implementing,  visualizing and  simulating  neural networks.   The   training  of the network  is  performed   by  presenting   180  values  of   encoder angles  at  the  one node  in the input layer   with  one  node  in the output layer    representing   the  corresponding  encoder error  in arc-min (i.e  $\theta_{ENC}$- $\theta_{TAB}$).  Since it  is well known  that, one hidden layer is sufficient  for  approximating  well behaved  functions [27],   we  have  also  used   one  hidden layer  in the present study.  The  activation function  chosen for  the present  study  is the  Sigmoid function.   
\par
Backpropapation  algorithm [28]  has been  by far,  the most popular and  widely used learning  technique  for  ANN training, we  have also used  the  same algorithm to  begin  with.   With regard to choosing  the  number of nodes  in the hidden layer  it is well  known that  while using too few nodes will  starve
the network of the resources that it needs to solve a particular problem,  choosing  too many  nodes  has  the risk of potential  overfitting   where the network tends to remember the training cases instead of generalizing the patterns. In order  to  find  the  optimum  number  of  nodes  in the  hidden  layer for  reproducing   the  error  profile  of a  particular  encoder  with  reasonably good accuracy,  we  changed the number of nodes in the hidden  layer  from  10 to 110 (in steps of 10 nodes)  and  the   corresponding  MSE  was  recorded at the end of ANN training   for  each  configuration.   The results of this  study  reveal that   while the  MSE   decreased   considerably   when  the number  of  nodes in the hidden  layer  was  increased from  20 to 80,
increasing  the  same  beyond  80  resulted  in  only  a  marginal   reduction  in the  MSE (at the cost of higher computation time).  It thus seems  that  one  hidden layer   with  $\sim$  80  nodes  is  quite  optimum for  reproducing   the  error  profile  of   the   encoder. 
Using  the  same  number  of  nodes  in the  hidden   layer  (i.e 80), we  then  studied  the  performance  of  several other  error minimization  algorithms  for  identifying   the  most suitable  algorithm  which yields  the  lowest  MSE.  The  algorithms  studied   were  the   Resilient  backpropagation,  Congugate gradient,  Fletcher Reeves,  Once Step Secant, Levenberg-Marquardt, Scaled conjugate  and  Quasi  Newton [29-31].   Amongst  all  these  algorithms,  it was  found  that  Levenberg-Marquardt  yields   the  lowest  MSE value of  $\sim$0.0083. Comparative MSE for other algorithms considered by us are given in Table 2 below.
%--------------------------------------------------------------------------------------
\begin{table}[h]
\begin{center}
\begin{tabular}{|c|c|}
\hline
$ALGORITHM$             &     $ MSE $     \\ 
\hline
$Res-Backpropagation$   &    $0.0150$    \\ 
\hline
$Conj- Gradient$         &    $0.0103$    \\
\hline
$Fletcher- Reeves$       &    $0.0144$   \\ 
\hline
$One Step-Secant$       &    $0.0147$   \\
\hline
$Levenberg-Marquardt$   &    $0.0083$    \\
\hline
$Scaled-Conjugate$       &    $0.0096$    \\
\hline
$Quasi- Newton$          &    $0.0091$    \\
\hline
\end{tabular}
\end{center}
\caption{\it{ Comparative MSE values for all the algorithms considered in  this  work.}} 
\end{table}
%--------------------------------------------------------------------------------------------
Fig.2a   depicts   the behaviour of  the   MSE as a function of  number of nodes in the hidden layer  for  the Backpropagation   and   Levenberg-Marquardt  algorithms. The  MSE  as a function of  number  of iterations  is shown  in Fig.2b   for  these  algorithms. 
%---------------------------------------------------------------------------------------------
%-------------------------Fig.2 --  ANN rms  error --------------------
\begin{figure}[h]
\centering
\includegraphics*[width=0.8\textwidth,angle=0,clip]{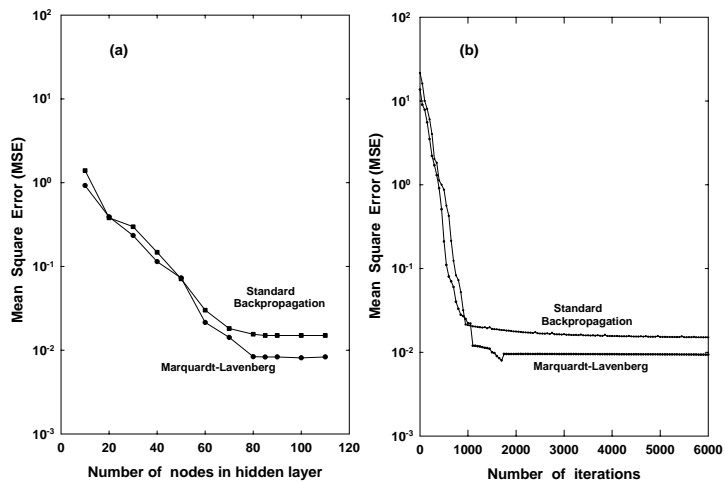}
\caption{\label{fig ---} (a)  Mean-squared  error  as a function of  number of nodes in the hidden layer. (b) Mean-squared error  as a function of number of iterations  for  80  nodes  in the  hidden layer.}   
\end{figure}
%--------------------------------------------------------------------------------------------------------
An examination  of  Fig. 2b  clearly  reveals  that   the  performance  of   the  Levenberg-Marquardt  algorithm  is  better  than  that of the   Backpropagation  algorithm.    It  is  worth mentioning  here  that  in   order  to  ensure  that  the  network   has  not  become  "over-trained" [32], the   ANN   training  is  stopped  when  the  normalised  rms   error   stops   decreasing   any further (somewhere  around 6000  iterations).  
The  MSE  for  the  ANN  configuration of 1:80:1  was   also   checked   for  the  remaining  3 encoders  and  it   was  found   that  it  varies  from  $\sim$  0.0098    to  $\sim$0.0165  for the  remaining  three  encoders. 
\par
A  detailed   study  for  determining   the  optimum  number  of  nodes  in  the  hidden  layer  in a  rigorous  manner  has  also been  conducted  to make  sure  that   the  ANN  configuration  chosen  is   not  more  complex  than  what is  actually  warranted.  While   several  empirical  results  have  been reported  which provide  adhoc,  heuristic  rules  for  selecting  the number of  nodes  in a  hidden layer,  we   have  followed   the   Singular Value Decomposition (SVD)  approach  for   finding  the  optimum  number  of  nodes  in  the  hidden  layer  by eliminating  the redundant  nodes [33].  Modification  of   the  ANN  structure  by  analyzing  how  much each  node  contributes to the actual output  of the neural network and  dropping  the  nodes  which do not  significantly affect  the output  is  also referred  to as  pruning. The  basic  principle  of  pruning    relies  on the fact  that if  two hidden nodes give the same outputs for  every input vector,   then   the  performance of the neural network  will not be   affected  by removing  one of  the nodes in the hidden layer. In the  SVD approach  redundant  hidden  nodes cause singularities in the weight matrix which can be identified  through  inspection of its  singular  values.   A non-zero  number of small singular values indicates redundancy in the initial choice  for the number of  hidden layer  nodes  and  the  approach  can be  safely  used  for  eliminating  these  nodes  to attain  the pruned  network model.   The  weight  matrix  (denoted by X in the present  work)  was  generated   by  finding  the  output  of  each  of  the 80  hidden nodes before  subjecting  them to  nonlinear transformation  (i.e  output  of  the  node  before  sigmoid function).  With   a total of 180 training patterns  and  one  hidden layer  with  80 nodes,   the matrix  X   thus   has  180 rows and 80 columns.   The  SVD  of the matrix  X is given by 
X=U S V $^T$, where   U and V  are the  orthogonal  matrices   and    S is a diagonal  matrix   with 80  rows and 80 columns.  The  matrix S  contains   the singular  values of  X  on its diagonal.     Plot  of  non-zero diagonal  elements   for   all the  4 encoders  is shown  in   Fig.3   and  
%---------------------------------------------------------------------------------------------------------------------------
%-----Fig.3 --  SVD  singular  values ----------------------------------------------------
\begin{figure}[h]
\centering
\includegraphics*[width=0.8\textwidth,angle=0,clip]{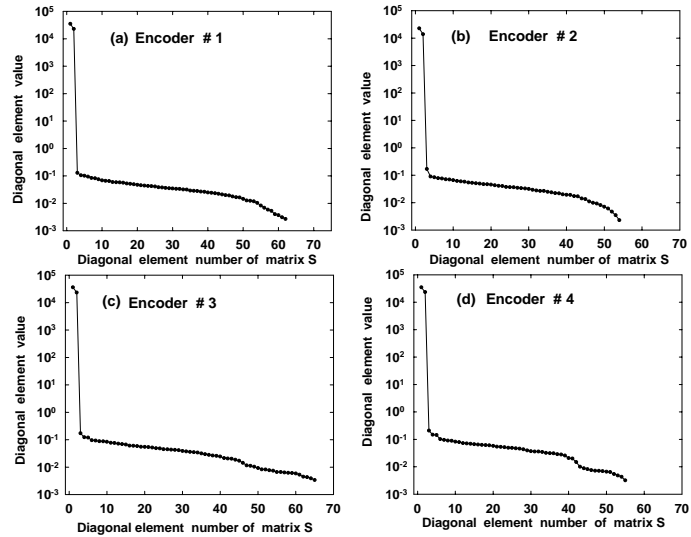}
\caption{\label{fig ---}(a-d)  The  singular  values  of the   matrix  X  as  a  function of   diagonal matrix  element  number.} 
\end{figure}
%---------------------------------------------------------------------------------------------
it is  evident  from  this  figure   that    number  of diagonal  elements  with  non-zero values   is  $<$80.    The   number of   diagonal  elements  with  non-zero values   for  the four  encoders   are    determined   to  be  the  following :
Encoder$\#1$  : $\sim$62; Encoder$\#2$ :   $\sim$ 54;  Encoder$\#3$   $\sim$65  and   Encoder$\#4$: $\sim$55.  The   above  analysis  clearly   suggests that  the  number  of  nodes  in the hidden layer  can be reduced  by using   the  SVD    method.   Using  the  optimized  number of nodes  in the hidden  layer   we  have  then   trained    the  network   again  so that  the   final  ANN  configuration  can be  tested. While it took ANN about 9 minutes to complete 10,000 training iterations on a P-IV machine (Intel(R)core, 2 Quad CPU, 2.39GHz, 3.25GB RAM) for 1:80:1 configuration, the training time was reduced to $\sim$ 4 minutes for a  SVD optimized typical configuration of 1:60:1.
 
%---------------------------------------------------------------------------------
\vskip 0.5cm
\noindent{\bf{ Testing  of Artificial Neural Network  and  results }}
\vskip 0.5cm
The ANN,  once  properly trained  with  optimized  number of nodes in the  hidden layer  for  each  encoder  error profile separately, was tested with respective  test   data files comprising  180 resolver angle values each (i.e.  1$^\circ$, 3$^\circ$, 5$^\circ$, 7$^\circ$, .......  , 359$^\circ$) which were not used during  training  of the network.  Denoting  the ANN-predicted  encoder error  as $\epsilon_{ANN}(\theta_{ENC})$,  we  then calculate  the residual  error (i.e  $\epsilon_{ANN}(\theta_{ENC})$-  $\epsilon_{obs}(\theta_{ENC})$ as a function of   $\theta_{ENC}$  for all 180  values  of  the encoder  angle.   Plots  of  the  residual  error  as a function  of   $\theta_{ENC}$ for the  four  encoders  are   shown  in Fig.4.  
%----------------------------------------------------------------------
%-----Fig.4 --  Error  profile  ANN --------------------------------------------------------------
\begin{figure}[h]
\centering
\includegraphics*[width=0.8\textwidth,angle=0,clip]{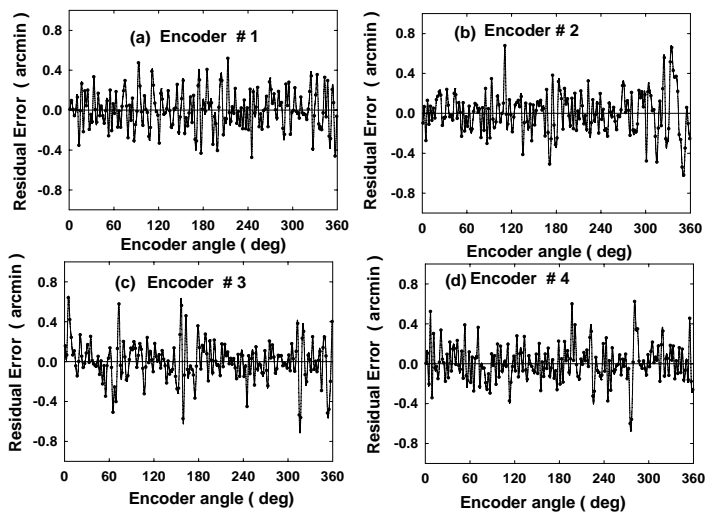}
\caption{\label{fig ---}(a-d) Residual error  ( i.e  $\epsilon_{ANNt}(\theta_{ENC}) $-$\epsilon_{obs}(\theta_{ENC})$) as a function  of   $\theta_{ENC}$ for  4 encoders. A smoothened  spline curve passing through the data points has also been  drawn  in these  figures so that a systematic structure (if any) present in the  residual error profiles  can  be  visualized  easily.} 
\end{figure}
%--------------------------------------------------------------------------------------
The  mean  absolute error  and the RMS  error  of the resulting  residual  error profiles 
(i.e  $\epsilon_{ANN}(\theta_{ENC})$-  $\epsilon_{obs}(\theta_{ENC})$  of the  4  encoders  are found to be following :  Encoder$\#1$  :  $\sim$0.15 arc-min and  $\sim$0.19 arc-min; Encoder$\#2$ : $\sim$0.16 arc-min and  $\sim$0.21 arc-min;  Encoder$\#3$   $\sim$0.14 arc-min and  $\sim$0.20 arc-min ;  Encoder$\#4$: $\sim$0.14 arc-min and  $\sim$0.19 arc-min. Comparison of these values with those presented in Fig.1 for the pre-compensation case indicates that   the   pre-compensation  values of mean absolute error, ranging between  $\sim$0.55 arc-min to  $\sim$1.77 arc-min  for  the 4 encoders, improve to  $\sim$ 0.15 arc-min  after compensation.   The  reduction  in  the  maximum residual  error  is   from  $\sim$$\pm$ 6 arc-min (pre-compensation) to    $\sim$$\pm$ 0.65  arc-min (post-compensation).
The  ANN-based error compensation procedure,  thus  involves  training  an ANN  appropriately  for each of the resolver-decoder combinations   by  using approximately  half of   the  calibration data  and  then  using  the  optimally trained   ANN-configuration  for  predicting   the  error  for any  encoder angle  in the range 0$^\circ$ to 360$^\circ$.   The  corresponding  corrected encoder  angle ($\theta_{COR}$)  is  determined   by  subtracting the  ANN-predicted error  from the measured  value of the encoder angle  ( i.e   $\theta_{COR}= \theta_{ENC}-\epsilon_{ANN}(\theta_{ENC})$). 
It is important  to  note here   that   the   overall  mean absolute error  and   the rms  error  values,   in  the  full   encoder  angle range 0$^\circ$ to 360$^\circ$  will   be  marginally   better  than  the  corresponding   test  data   because of the  fact that  the  ANN  will  always  yield  better results  around  encoder angle  values   which were  used  during  training (i.e.  0$^\circ$, 2$^\circ$, 4$^\circ$, 6$^\circ$, .......  , 358$^\circ$).  
\par
In order  to compare  the    results  of  the   ANN-based  error  compensation procedure  with  those  obtained  from the Fourier series-based    approach,  we will   first   briefly  present  the  results  of the later  method.  Assuming   that  the  error profiles of the encoders can be  approximated  by a Fourier series representation, as already  demonstrated  in our   earlier  work [7], a  general expression for the  same  can be written in the following manner.
%---------------------------------------------------------------------------------
\begin{equation}
\epsilon_{fit}(\theta_{ENC})= a_0+  \sum \limits_{n=1}^N 
 [ a_n  \;  \cos(n \;\; \theta_{ENC}) +b_n \;  \sin(n \;\; \theta_{ENC}) ]
\end{equation}
%-------------------------------------------------------------------------------------
where  $\epsilon_{fit}(\theta_{ENC})$ is  the  fit to  $\epsilon_{obs}(\theta_{ENC}$) values,  $\epsilon_{obs}(\theta_{ENC}$)  is  the  observed error in the encoder angle  (i.e $\epsilon_{obs}(\theta_{ENC}$)= $\theta_{ENC}$- $\theta_{TAB}$),  N  is an integer  indicating the  maximum  order of the harmonic  which needs  to be considered,  $a_{0}$ is a constant signifying the amplitude of the  dc component, $a_n$ and $b_n$  are  the even and odd Fourier coefficients, respectively. However, before  approximating  the  error profile of an  encoder  with  a   Fourier series with  manageable number of terms,  it  becomes important   to first perform  a detailed harmonic analysis of each  error profile   so that one can  retain only   those  terms  in  the expansion  which  are  of  significant  amplitude.  The most prominent 10  harmonics  (arranged in descending  order of their  amplitudes)  which have  been  used in the Fourier series expansion   for the  4  error profiles  shown in  Fig.1   are the following: Encoder $\#$1:  (1, 2, 16, 0, 14, 48, 4, 32, 3, 84);   Encoder$\#$2 : (0, 1, 2, 16, 14, 32, 48, 10, 6, 4); Encoder $\#$3:  (1, 16, 2, 48, 14, 12, 32, 0, 6, 10) and  Encoder$\#$4:  
(0, 1, 16, 2, 32, 12, 4, 3, 14, 48).  
\par
To  summarize  the quantum of the resulting  improvement when software-based error compensation is used, we have  given in  Table 3, the  mean absolute error and RMS error before and after compensation  for   all the four encoders. 
%---------------------------------------------------------------------------------------
\begin{table}[h]
\begin{center}
\begin{tabular}{|c|c|c|c|}
\hline
$ENCODER$     &   $ Before$        &       $Post Compensation$  &  $Post Compensation$ \\ 
$ No. $       &   $Compensation$   &              $Fourier$     &     $ ANN $ \\ 
$      $      &   $ MAE/RMS$       &              $MAE/RMS$     &     $ MAE/RMS $ \\ 
$      $      &   $ (arc-min)$     &              $(arc-min)$   &     $ (arc-min) $ \\ 
\hline
$1 $ & $1.33/1.48$ & $0.16/0.21$ & $0.15/0.19$ \\ 
\hline
$2$& $0.55/0.65 $ &  $0.16/0.20 $ & $0.16/0.21 $ \\ 
\hline
$3$  & $1.09/1.35 $ &  $0.14/0.18 $ & $0.14/0.20 $ \\     
\hline
$4 $  &  $1.78/1.31 $ &  $0.16/0.21 $ & $0.14/0.19 $ \\
\hline
\end{tabular}
\end{center}
\caption{ \it{ Comparison  of   Mean Absolute Error (MAE) and the RMS error  for   pre-compensation and    post-compensation  cases. } }
\end{table}
%-----------------------------------------------------------------------------

On comparing  the performance  of  the  ANN-based  error  compensation procedure  with  that  of the  Fourier series-based approach,
it  is evident from  Table 3  that  the  two  methods  yield  almost  similar  results. However the main advantage of the ANN-based compensation is that it can be applied to a wide range of sensors with any arbitrary error profile. 
Reduction of the  maximum residual  error  from  $\sim$$\pm$ 6 arc-min (pre-compensation) to    $\sim$ $\pm$ 0.65  arc-min (post-compensation), alongwith 
significant  reduction  in the  mean  absolute error  and the RMS  error   of the residual error profiles  clearly  illustrates   that  the  accuracy of low-cost, resolver-based  encoders can be  improved significantly  by employing a  suitable  software-based error compensation procedure. The  mean absolute error  and RMS error  of the  residual error  profiles   after applying  the   ANN-based error  compensation  method are  found out  to be  in the range  $\sim$ 0.14  arc-min to $\sim$ 0.16  arc-min  and   $\sim$ 0.19  arc-min  to  $\sim$0.21 arc-min,  respectively. 
It  is  worth mentioning  here  that the  use  of   a dedicated  ANN software package  is  necessary   only  during   the  training  of  the  ANN.  
Once satisfactory training of the ANN is  achieved, the corresponding ANN generated weight-file can  be  easily used  by  an  appropriate  subroutine  of the  data acquisition  program  for  determining    the  corrected  encoder angle.  We have  also successfully  implemented 
the ANN-based  error compensation  procedure in our  data  acquisition  program,  by directly using the ANN generated weight-file,  so that the corrected encoder angle  can  be predicted directly without using the ANN software package. An added advantage of using the weight-file directly is that the corrected encoder angle information can be made available 'on-line' if required.
%------------------------------------------------------------------------------------------

\vskip 0.5cm
\noindent{\bf{Discussion}}
\vskip 0.5cm
With  regard  to  other  compensation  methods,   while as  a  number  of   methods   have  been proposed for  improving   the  accuracy of  resolver-based  encoders,  we  will only   present  here  a brief  summary  of  some  of the  relevant  published results.  Assuming   that factors like amplitude  imbalance, quadrature error, inductive harmonics  and   excitation  signal  distortion  are  mainly  responsible  for   making   the output signals  of a  resolver  deviate  from    ideality,  it has    been  shown in  [5]  that  by   using  appropriate signal  processing,  quadrature error   can  be  eliminated.   It  has  also been pointed  out in  [5]  that   all even harmonics  in the resolver  signals  can  also  be  cancelled,    if the resolver  is constructed  with   complementary phases. 
A  compensation circuit    for    decreasing    the  quadrature  error and the  amplitude  imbalance  of  a resolver  is   reported  in  [34].  The results  of  their  study,   reported  for   a 34-pole  resolver, indicate   that  the     accuracy  can be  improved  from  $\sim$$\pm$15 arc-min  to $\sim$$\pm$2 arc-min  by   using   the  compensation circuit.  As a  replacement   
 for    the  conventional  phase-lock   loop  method ,   a  resolver-to-dc converter circuit   based  on  
linearization  approach  has  been proposed  
in  [35],  but  an  accuracy   of  only  $\sim$12 arc-min  has  been achieved  by  using  this   method.   A  gain-phase-offset  correction  method  using  cross-correlation   has   been   used  in  [36]   for suppressing   systematic  errors of resolvers  and optical encoders.   However,  it  is  also pointed out  in [36]  that   the method is not  suitable  for  fast point-to-point   positioning 
 with  small distances  (i.e.   with   less than  one period  of  the  resolver  line  signals). 
More recently,    in   an attempt to  overcome   the limitation  of   the   cross-correlation  method, 
 an  adaptive  phase-lock loop   based   R/D converter  has  also been  proposed   in    [37]  for   estimating   the  angular  position  and  speed  of   resolver sensor   systems.     Using     a DSP  based  system,   the authors   [37] claim  that    in  addition    to   compensating   for   amplitude imbalance,  quadrature error    and  harmonic  
distortion,  there    is  also  a   saving  in  hardware. 
A  novel ANN based method  has   been   used     for adaptive online correction and interpolation of encoder signals [38].   The method followed by them uses a two stage RBF based ANN model,   where  the first stage of the network is used for correction of incoming non-ideal encoder signals  and    the second stage is used to derive higher order sinusoids from corrected signals of the first stage.  Although the method used by them  gives similar results compared to the lookup table method,    it has the added  advantage  with respect to the memory storage requirements, since when the number of data points calibrated in a three-dimensional workspace increases by a factor of N, the number of  entries in the look up table will increase by N$^3$.  A wavelet-based  algorithm [39]   has been  used  for extracting  the  integral and differential nonlinearity  of encoders.
\par
On  the  basis  on the  above  discussion  
(and  other  published  literature   on the subject)  one  can  safely  say   that, 
in addition  to  hardware  improvements  of the encoder system,   there   is  also a strong  need  for  developing    software-based  compensation   methods  for 
achieving    accuracies of   $\lesssim$  1 arc-min.    Significant   improvement    in the  accuracy,    from  $\sim$$\pm$6arc-min  to   $\sim$$\pm$0.65 arc-min,  achieved  in   the present  work,     clearly demonstrates  that  ANN-based  compensation   method   can   indeed  help  in   improving  the   accuracy   of   low-cost  encoders.  Similar  error  reduction  for  different  applications, by  using ANN-based  compensation  method,  have  also  been  reported  by  others  [9,16,19].
%--------------------------------------------------------------------------------------------

\vskip 0.5cm
\noindent{\bf{Conclusions}}
\vskip 0.5cm
An ANN-based  error compensation procedure has been developed in this work to improve the accuracy of low-cost  resolver-based  16 bit encoders.  The  procedure   has  allowed  us  to  use the existing resolvers  at an accuracy  which is within the limits of the encoder resolution.   Reduction of the  maximum residual  error  from  $\sim$$\pm$ 6 arc-min (pre-compensation) to    $\sim$$\pm$ 0.65  arc-min (post-compensation)
clearly  illustrates  that  the  accuracy of  low-cost, resolver-based  encoders can be  improved significantly   by employing an  ANN-based error compensation procedure.
This procedure   has  been   implemented   for 4  encoders by  training 
the  ANN  with  their  respective  error profiles   and  then  using  their  corresponding   ANN-generated   weight  files  for  determining  the  corrected  encoder angle. 
We believe  that a little  expense involved in  generating the calibration data   is  quite  justifiable   keeping  in  mind  that using optical encoders  of comparable accuracy,  in  lieu  of  resolver-based  encoders,  is a more costlier option[1].  Because of their delicate nature, the optical encoders are very sensitive to hostile environment (temperature, moisture, dust etc.) and  are  prone to reliability problems in application where the  sensors  need  to be   installed away from the control room. Furthermore,  since  our   application   needs  the encoder angle data to be transferred over long cables ($\sim$~ 75 m), use  of a  suitable driver also  involves  an additional cost.  On the other hand,    resolver-based   encoders can transmit  analog angle data over much longer lengths of cable, besides having a very wide working temperature range.
%--------------------------------------------------------------------
%-------------------------------------------------------------------------------------
\vskip 0.5cm
\noindent{\bf{Acknowledgements}}
\vskip 0.5cm
The authors would to like to thank all the concerned  colleagues of the  Centre for Design and Manufacture  of BARC  for their  help in  calibrating  the encoders  using  the  Rotary Table setup  of  the Coordinate Measuring Machine. We would also like to thank the anonymous referees  and  the adjudicators  for making several helpful suggestions. 
%----------------------------------------------------------------------------------------------- 
%----------------------------------------------------------------------------------------------- 

%\noindent*{References}
\par


\begin{thebibliography}{10}

\bibitem{book1} Gasking J (Eds) 1980 {\it  Synchro and Resolver Conversion, (http://www.naii.com)} 

\bibitem{book1} Moog Components Group  2004  {\it  Synchro and Resolver Engineering Handbook, (http://www.moog.com)} 

\bibitem{book1} Gasking J (Eds) {\it Analog Devices Application Note, AN-263,} \newline
{(www.analog.com/UpdatedFiles/ApplicationNotes/394309286AN263.pdf)} 

\bibitem{} Hanselman D  1990 Resolver Signal Requirements for High Accuracy Resolver-to-Digital conversion  {\it  IEEE Trans. Ind. Electronics}
{\bf 37} 556-561 
 
\bibitem{} Hanselman D  1991 Techniques for Improving Resolver-to-Digital Conversion Accuracy {\it IEEE Trans. Ind. Electronics}
{\bf 38} 501-504  

\bibitem{} Kaul S K  et al 1997   Use of  Look Up table  improves  the accuracy of Low -cost  resolver-based  absolute  shaft encoders  {\it Meas. Sci.  Technol.} {\bf 8} 329-331  

\bibitem{} Kaul S K  et al 2008   Improving  the  accuracy of low-cost  resolver-based  encoders  using  harmonic  analysis   {\it  Nucl. Instr. and Meth. A } {\bf 586} 345-355

\bibitem{} Wu S M et al 1989  Precision machining without precise machinery  {\it Annals of CIRP } {\bf 38 (1)} 533-536

\bibitem{} Chen J S  1996 Neural Network-based Modelling and Error compensation of thermally-induced Spindle Errors {\it Int. Jour. Adv. Manuf Technol} {\bf 12} 303-308 

\bibitem{} Tai H M et al 1992 A Neural Network-Based Tracking Control System {\it IEEE Trans. Ind. Electron. } {\bf 39} 504-510

\bibitem{} Ozaki T et al 1991 Trajectory control of robotic manipulators using neural networks {\it IEEE Trans. Ind. Electron. } {\bf 38} 195-202

\bibitem{} Kraft L G et al 1990 A comparison between CMAC neural networks control and two traditional adaptive control systems {\it IEEE Control Systems Mag. } {\bf 10 (3)} 36-43

\bibitem{} Psaltis D et al 1988 A multilayer neural network controller {\it IEEE Control Systems Mag. } {\bf 8 (2)} 17-21

\bibitem{} Yamada T et al 1990 An extension of neural network direct controller {\it Proc. Int. Workshop Intelligent Robots Syst }  619-626

\bibitem{} Kawato M et al 1987 A hierarchical neural network model for control and learning of voluntary movement  {\it Biol. Cybern} {\bf 57} 169-185

\bibitem{} Barto A et al 1983 Neuron like adaptive element that can solve difficult learning control problems {\it IEEE Trans. Syst. Man. Cybern.} {\bf SMC-13} 834-846

\bibitem{} Guo H J 2003 Sensorless Driving Method of Permanent-Magnet Synchronous Motors Based Neural Networks {\it IEEE Trans. Magnetics} {\bf 39} 3247-3249

\bibitem{} Ziegert J C et al 1994 Error compensation in amchine tools : a neural network approach {\it Jour. Intelligent  Manufacturing} {\bf 5} 143-151

\bibitem{} Tan K K et al 2006 Geometrical Error Modelling and Compensation Using Neural Networks {\it IEEE Trans. Syst. Man. Cybern.} {\bf 36} 797-809

\bibitem{} Computer Conversions Corporation Catalog  1989 , {\bf CAT. 89 } 

\bibitem{} Koul R  et al 2007   The  TACTIC atmospheric  Cherenkov imaging telescope {\it  Nucl. Instr. and Meth. A } {\bf 578} 548-564

\bibitem{} Tickoo A K  et al 1999   Drive control system for the  TACTIC  gamma-ray telescope  {\it  Exp. Astron. } {\bf 9} 81-101

\bibitem{} Coordinate Measuring Machine Documentation, Zeiss   1981   {\it http://www.zeissmetrology.com} 

\bibitem{} Tagliaferri R et al 2003  Neural Networks in Astronomy. {\it Neural Networks.} {\bf 16} 297-319.        

\bibitem{} Dhar V K  et al 2009 ANN based energy reconstruction procedure for TACTIC $\gamma$-ray telescope and its comparison with other conventional methods. {\it Nucl. Instr. and Meth. A } {\bf 606} 795-805.

\bibitem{} http://www.mathworks.com

\bibitem{} White  H  1989   Learning  in artificial neural networks: a statistical perspective {\it  Neural Computation  } {\bf 1} 425-464

\bibitem{} Rumelhart D E et al 1986 Learning representations by back-propagating errors {\it Nature.} {\bf 323} 533-536.

\bibitem{} Hagan M T et al 2003 Neural Network Design, Thomson Asia Pvt Ltd, Singapore, Vikas Publishing Pvt Ltd, N Delhi.

\bibitem{} Zurada J M 2006 Introduction to Artificial Neural Systems, Jaico Publishing   House, Mumbai.

\bibitem{} Press W H et al 1998 Numerical Recipes 2nd Edition, Cambridge University Press.

\bibitem{} Duda R O et al 2002  Pattern Classification, John Wiley And Sons, New York.

\bibitem{} Kanjilal P P  et al   1993   Reduced-Size  neural networks through singular value decomposition and subset selection
 {\it Electronics  Letters} {\bf 29} 1516-1518

 \bibitem{} Mingji Liu  et al 1999  Error  analysis and compensation of multipole resolvers  {\it Meas. Sci.  Technol.}
{\bf 10} 1292-1295  
 
\bibitem{} Benammar M   et al 2004  A novel resolver-to-360$^0$ linearized converter 
{\it IEEE  Sensors Journal.} {\bf 4} 96-101 

\bibitem{} Bunte A  and  Beineke S  2004   High performance   speed  measurement   by suppression of systematic  resolver and  encoder errors {\it IEEE Trans. Ind. Electron. } {\bf 51} 49-53

\bibitem{} Sarma S  and  Venkateswaralu  A  2009   Systematic  error  cancellations  and  fault  detection of resolver  angular  sensors using  a  DSP based  system  {\it To  appear  in  Mechatronics.} {\bf  doi: 10.1016/j.mechatronics.2009.09.002}

\bibitem{} Tan K K et al 2005 Adaptive online correction and Interpolation of Quadrature Encoder Signals Using Radial basis Functions {\it IEEE Trans. Control Systems Technol. } {\bf 13} 370-376

\bibitem{}  Pereira J M Dias   et al  2007   Wavelet techniques: A suitable tool  to characterise and optimize encoders'  based systems 
{\it  Measurement  } {\bf 40} 264-271

 
%------------------------------------------------------------------------------------------------ 
\end{thebibliography}
\end{document}